\documentclass[a4paper, 10pt, conference]{IEEEtran} 

\usepackage{graphicx}
\usepackage{amsmath}
\usepackage{amssymb,tikz}
\usepackage{graphicx}
\usepackage[most]{tcolorbox}
\usepackage{float}
\usepackage[nodayofweek]{datetime}
\newdateformat{mydate}{\twodigit{\THEDAY}{-}\shortmonthname[\THEMONTH]{-}\THEYEAR}
\renewcommand{\date}{16-Apr-2020}
\usepackage{xcolor,multirow}
\usepackage{diagbox}
\usepackage{rotating}
\usepackage{tikz} 
\usepackage{xltabular}
\usepackage{tabu, booktabs}
\usepackage{makecell}
\usepackage{array}

\newcommand\Tstrut{\rule{0pt}{4ex}}         
\newcommand\Bstrut{\rule[-7ex]{2pt}{2pt}}   

\newcommand\TstrutNorm{\rule{0pt}{2.6ex}}         
\newcommand\BstrutNorm{\rule[-1ex]{0pt}{0pt}}   
\usepackage{cite}
\usepackage{amsmath,amssymb,amsfonts}
\usepackage{algorithmic}
\usepackage{graphicx}
\usepackage{textcomp}
\usepackage{xcolor}
\def\BibTeX{{\rm B\kern-.05em{\sc i\kern-.025em b}\kern-.08em
    T\kern-.1667em\lower.7ex\hbox{E}\kern-.125emX}}
\begin{document}

\title{Unsupervised Action Localization Crop in Video Retargeting for 3D ConvNets}

\author{\IEEEauthorblockN{Prithwish Jana}
\IEEEauthorblockA{\textit{Dept. of Computer Science \& Engineering} \\
\textit{Indian Institute of Technology Kharagpur}\\
Kharagpur, India \\
pjana@ieee.org}
\and
\IEEEauthorblockN{Swarnabja Bhaumik}
\IEEEauthorblockA{
\textit{Larsen \& Toubro Infotech Ltd.}\\
Ghaziabad, India \\
swarnabjazq22@gmail.com}
\and
\IEEEauthorblockN{Partha Pratim Mohanta}
\IEEEauthorblockA{\textit{Electronics \& Communication Sc. Unit} \\
\textit{Indian Statistical Institute}\\
Kolkata, India \\
partha.p.mohanta@gmail.com}
}

\maketitle

\begin{abstract}
Untrimmed videos on social media or those captured by robots and surveillance cameras are of varied aspect ratios. However, 3D CNNs usually require as input a square-shaped video, whose spatial dimension is smaller than the original. Random- or center-cropping may leave out the video's subject altogether. To address this, we propose an unsupervised video cropping approach by shaping this as a retargeting and video-to-video synthesis problem. The synthesized video maintains a 1:1 aspect ratio, is smaller in size and is targeted at video-subject(s) throughout the entire duration. First, action localization is performed on each frame by identifying patches with homogeneous motion patterns. Thus, a single salient patch is pinpointed per frame. But to avoid viewpoint jitters and flickering, any inter-frame scale or position changes among the patches should be performed gradually over time. This issue is addressed with a polyB\'ezier fitting in 3D space that passes through some chosen pivot timestamps and whose shape is influenced by the in-between control timestamps. To corroborate the effectiveness of the proposed method, we evaluate the video classification task by comparing our dynamic cropping technique with random cropping on three benchmark datasets, viz. UCF-101, HMDB-51 and ActivityNet v1.3. The clip and top-1 accuracy for video classification after our cropping, outperform 3D CNN performances for same-sized random-crop inputs, also surpassing some larger random-crop sizes.
\end{abstract}


\section{Introduction}
Visual information on the Internet is accessed through a multitude of display devices with varied screen dimensions and resolutions. This calls for \textit{video retargeting}~\cite{liu2006video} whereby a video is adapted to view with clarity, on a display different from what was originally intended. The scenario is somewhat similar when it comes to training 3D Convolutional Neural Networks (CNNs) with videos, for different tasks in computer vision e.g. activity recognition, localization, detection, etc. Videos regularly encountered in day-to-day life are typically elongated rectangular in shape with aspect-ratio like 16:9, 4:3 etc. However, conventional 3D ConvNets e.g. 3D-ResNet~\cite{kataoka2020would}, (2+1)D-ResNet~\cite{tran2018closer}, I3D~\cite{carreira2017quo}, 3D-ResNeXt~\cite{xie2017aggregated} typically accept square-shaped video frames as input. Practitioners prefer sizes within $224\times224$, if not less. Larger input size could require more training samples and deeper networks, implying increased computation time and space. Simple linear down-scaling of the frames or random cropping can render the objects too small to be identifiable or remove valuable content. Further, real-world computer vision tasks aim to tackle unconstrained videos that are captured by amateurs and have negligible quality controls. As a consequence, the biggest challenge of working with such videos is that the cropping algorithm should efficiently cope up with the myriad of quality artifacts present in such videos e.g. inferior lighting conditions, jittery camera movements and variations in view-point.

In this paper, we design a lightweight dynamic video cropping technique that utilizes unsupervised action localization in frames and guarantees temporal cohesiveness amongst the cropped action-patches obtained from successive frames. The remainder of this paper is organized as follows. We discuss some of the earlier works on video cropping and retargeting in Section~\ref{relWork}. In Section~\ref{propMethod}, we elaborate on the proposed method. In Section~\ref{exptSec}, we discuss the implementation, datasets used and experimental results. Finally, we conclude with an epilogue and future scopes of improvement in Section~\ref{conclusion}.

\begin{figure*}[!h]
	\centering
	\includegraphics[width=0.92\textwidth]{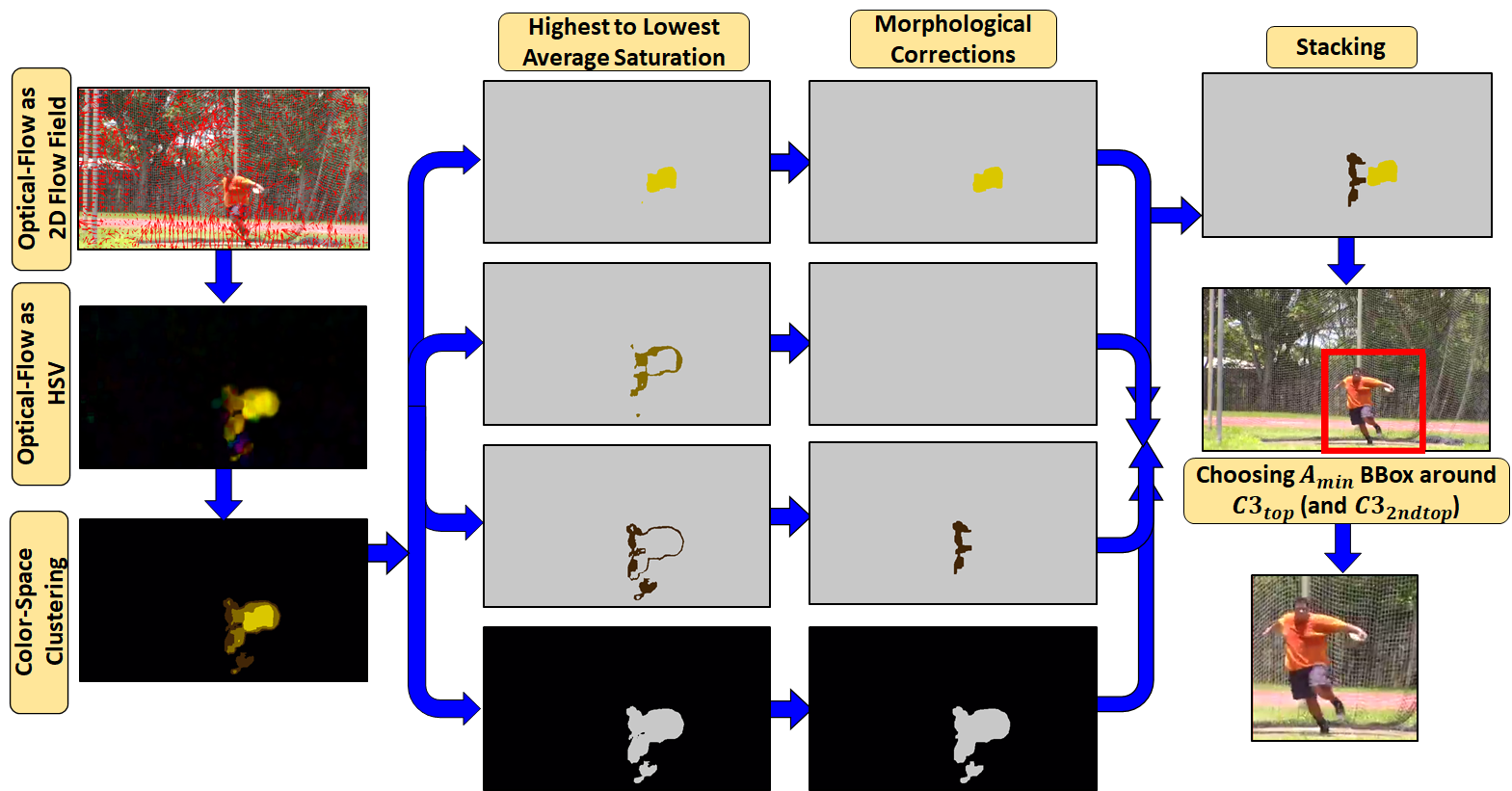}
	\caption{Localization of action in a single frame of a video on Discus Throw}
	\label{fig:AxnLocIndivFrames}
\end{figure*}

\section{Related Work}
\label{relWork}

Karpathy et al.~\cite{KarpathyCVPR14} posited about the issues of cropping and re-scaling to a fixed dimension, in presence of high temporal variance. They introduced multiresolution CNN that inputs low-resolution resized frame along with a center crop. But as Wang et al.~\cite{wang2010motion} argues, some objects can cover majority of frame and may not be centrally-placed. As such, an upper-limit of crop-size and fixed center coordinates need to be relinquished. Most of the video retargeting methods~\cite{li2020perceptual} apply image retargeting techniques on individual frames. In recent times, a majority of per-frame spatial action localization methods~\cite{kalogeiton2017action} in untrimmed videos, employ object detection through R-CNNs. As this adds an overhead of large end-to-end training time~\cite{bhaumik2021event}, researchers prefer lightweight unsupervised approaches. Patches of significant inter-frame motion can be identified by binarization~\cite{jana2017fuzzy} on the magnitude component of optical-flow matrix among consecutive frames. Incorporating image-based pixel-wise semantic segmentation~\cite{mukherjee2020two} on the RGB frames corresponding to such patches can bring out salient objects that exhibit significant inter-frame motion. But this is also overly computation intensive. Moreover, such independent image retargeting schemes that do not consider temporal constraints, may lead to temporally incoherent results. Temporal constraint may be categorized into \textit{local} or \textit{global}. Li et al.~\cite{li2018faster} do not restrict seam movement within frames (local constraint) to allow discontinuous seam carving. Liu et al.~\cite{liu2017fastshrinkage} used aggregation-based ConvNet to extract features from human gaze shifting trajectory. Thus, while local constraint is imposed on a sliding window of consecutive frames, global methods consider the overall video thereby proffering better~\cite{li2020perceptual} temporal consistency.  

For cropping videos for 3D ConvNets, different crop aspect-ratios for different frames is not beneficial because it distorts optical flow. Further, when temporal constraints are ignored, preservation of objects' dimension in consecutive frames and retention of motion patterns, cannot be guaranteed simultaneously. To address this problem of video cropping, we crop high-action patches from frames of video and connect them together in such a way that the inter-frame change in dimension and position is gradual. Further, the original video's subject is targeted in the cropped video irrespective of its relative position within a frame -- whether it is present at the frame's center, at its boundaries or somewhere in between. 

\section{Proposed Methodology}
\label{propMethod}

Firstly, we detect a single square-shaped patch of focused action in each frame of an untrimmed video. Next, we bring about temporal consistency between these patches by adjusting position and dimension through curve fitting. These patches when stitched together, proffers a video that is an intelligently spatially-cropped version of the original, specifically
focused towards the video-subject throughout the whole video duration.

\begin{figure*}[!ht]
	\centering
	\includegraphics[width=0.95\textwidth]{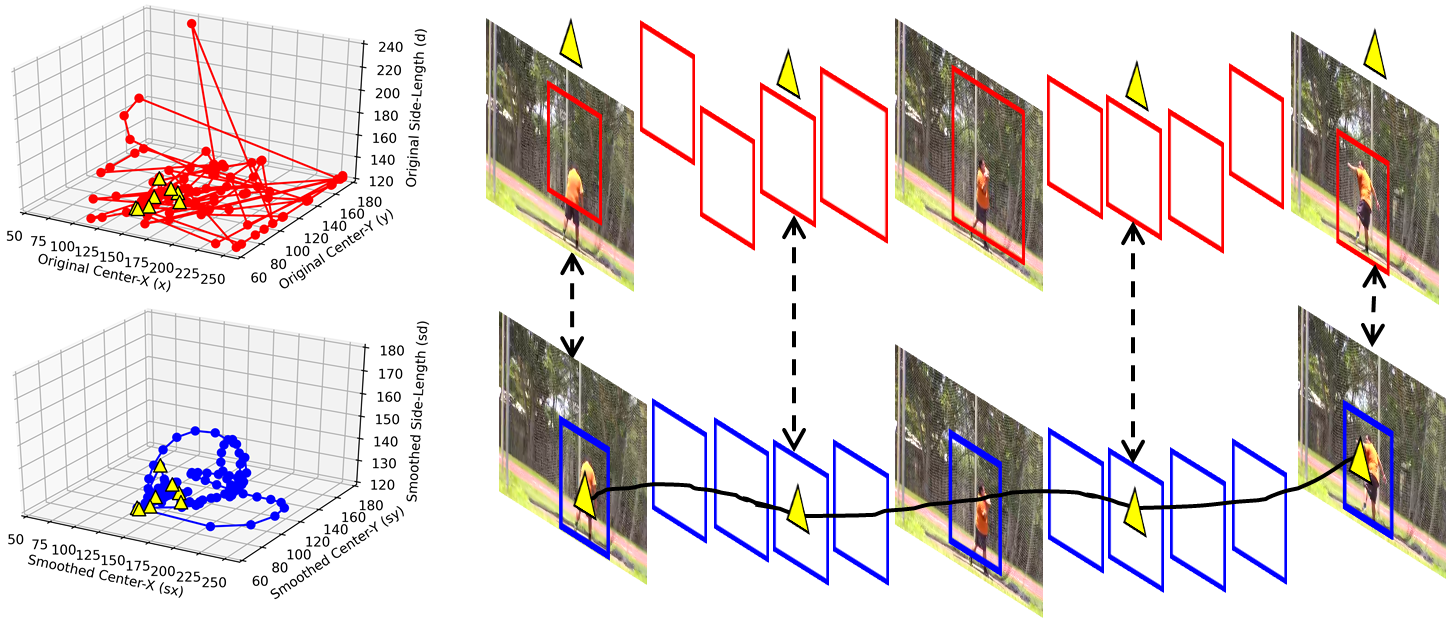}
	\caption{Position of 3D points $C_i=(x_i, y_i, d_i)$, \textit{before} (top row) and \textit{after} (bottom row) applying the proposed temporal consistency. On the top-left, successive red dots denote control points used for polyB\'ezier between two consecutive pivots (yellow triangles). On the bottom-left, blue dots are the corresponding interpolated points after proposed curve-fitting. Smoothened curvature indicates gradual inter-frame size/position changes for action patches.}
	\label{fig:bezierInterpolation}
\end{figure*}

\subsection{Unsupervised Action-Localization in Individual Frames}
\label{axnLoc}

We intend to localize a single square-shaped patch of focused action in every single video frame. From one frame to its immediate next, motion of video objects can be quantified by computing dense optical-flow~\cite{farneback2003two}. This motion can be visualized by representing the dense 2D flow field as HSV image, where H and S channels respectively correspond to \textit{orientation} and \textit{magnitude} of flow vectors. Color-space clustering through K-means++ brings under same label, `motionally'-related pixels bearing resemblance with respect to speed and direction of motion. 

We observed many instances of clusters where, far-apart large spatial blobs are frequently inter-connected by narrow pathways with pixels from the same cluster. To identify objects from the clusters, these far-apart blobs should be disconnected. Thus, after separating out each cluster into an individual image, we perform morphological closing and opening operations followed by removal of noisy components. To retrieve back a single HSV, each processed image pertaining to a cluster is stacked one upon another. But we maintain an order in this regard. Starting with the image corresponding to cluster with least average saturation, we successively go on stacking with increasing values of average saturation. This is because, average saturation of all pixels in a cluster gives an idea of the motion attached to it -- more the motion, more the chance of it containing the subject of the frame. Effectively, we can allow a layer with less motion to be overshadowed by a layer that is positioned higher on the motion scale. But the converse is disallowed as detail preservation is more important for latter. 

Finally, we identify a patch of the image that has high chance of representing the subject of a frame. After the previous step, each cluster may get divided into multiple spatially separated patches (if not already separated from before). From these, we identify the connected components all of whose pixels fully belong to one single cluster, henceforth called as Cluster Connected Components (C3). Disregarding C3s that touch the border of the image, we identify the top two C3s having the highest average saturation (referred to as $C3_{top}$ and $C3_{2ndtop}$). Here, border-touching C3s are not considered because of the spatial coherence assumption of optical-flow algorithms. At the image border, pixels with motion meet stationary pixels and this assumption is not satisfied leading to exaggeration of motion through false optical-flow representations. If the bounding-box of $C3_{top}$ covers a minimum acceptable area ($A_{min}$) of the image deemed suitable for representing the frame's subject, then this is the only chosen component. If not, we expand $C3_{top}$'s bounding box equally from all sides until its area exceeds $A_{min}$. If this expanded bounding box overlaps with the bounding-box of $C3_{2ndtop}$, we choose both $C3_{top}$ and $C3_{2ndtop}$. When both the boxes overlap, it is highly possible that they together may correspond to something more semantically meaningful. Else if there is no overlap, we choose only $C3_{top}$. 

Be it only $C3_{top}$ or $C3_{top}$ and $C3_{2ndtop}$ combined, we consider the bounding-box around the chosen C3(s). This is made to assume shape of a square by incrementing the lower dimension (length or width). This square patch is the most motion-heavy semantically-meaningful patch in a video frame. Except for some minute fraction of frames (which we will redress in Section~\ref{tempConst}), this unsupervised approach is successful in identifying the \textit{subject} in a majority of the frames. The overall sequence of steps is elucidated in Figure~\ref{fig:AxnLocIndivFrames}.

\subsection{Temporal Consistency for Localized Actions}
\label{tempConst}
When the subject of a video (or any object in particular) is observed across successive video frames, its space-time displacement is meagre. As such, the action localization bounding-boxes should also change minimally from frame to frame, implying substantial overlap of the box proposals for consecutive frames. But simply choosing the action patches of each video frame (from Section~\ref{axnLoc}) may give rise to temporally inconsistent outcomes (viewpoint jitters and flickering artifacts), as illustrated in the top row of Figure~\ref{fig:bezierInterpolation}. Thus, some adjustments are required so as to obtain a smooth \textit{space-time action-localized tube} spanning the entire video duration. We represent the square patch of the $i^{\text{th}}$ frame ($i=0,1,\cdots ,F-1$) by a 4D coordinate $(x_i,y_i,d_i,i)$ where $x_i,y_i$ is the spatial co-ordinate of the square center, $d_i$ is the side-length of the square and $i$ is the frame's timestamp. One way of achieving smoothness is to fit a $(F-1)$-degree B\'ezier curve $B(t)$ from $C_0=(x_0,y_0,d_0)$ through $C_{F-1}=(x_{F-1},y_{F-1},d_{F-1})$. The smoothed square patch will thus be $(sx_i, sy_i, sd_i, i)$ where $(sx_i, sy_i, sd_i)=B(\frac{i}{F-1})$. But this tends to be erroneous as the curve is only assured to start and end at $C_0$ and $C_{F-1}$ respectively by the endpoint interpolation property of B\'ezier curves. Intermediate points are treated equally whether temporally consistent or not. As such, not only the inconsistent patches, but also those that are already temporally consistent are at the risk of being displaced in influence of its temporally inconsistent neighbors.

\subsubsection{Pivot Selection for polyB\'ezier Endpoints}
\label{pivotSelect}

Instead of a single curve, we propose a piecewise B\'ezier curve or a \textit{polyB\'ezier} with certain \textit{pivot points} (through which the composite curve will pass). For this, we construct a set $PT\subset \{0,1,\cdots ,F-1\}$ that will serve as the collection of pivot timestamps. To state it otherwise, $C_t$ will be considered as a pivot point \emph{iff} $t\in PT$. Rest of the points will behave as control points like before. To construct $PT$, we formulate a scoring mechanism to quantify the temporal cohesiveness of each existing box proposal $C_i$ as follows:

\begin{equation}
	\label{eq:scoreProposals}
	S_i = \sum_{f=0,f\neq i}^{F-1}\frac{IoU_{i,f}}{\left| i-f \right| }\;,\;\;\;\;\;\; i=0,1,\cdots ,F-1
\end{equation}

In Eqn.~\ref{eq:scoreProposals}, $IoU_{i,f}$ denotes the Intersection-over-Union (IoU) ratio for proposals from $i^{\text{th}}$ and $f^{\text{th}}$ frame. $\left| i-f \right| $ is the timestamp difference amongst those two frames. This score is high for those proposals which bear a high IoU ratio with temporally nearby frames -- effectively implying temporal consistency. Thereby, we follow an iterative pivot-selection procedure. In each iteration, we choose that $i$ as the pivot timestamp which corresponds to highest score. And we exclude four timestamps from consideration in successive iterations, viz. two timestamps before (${i-2}, {i-1}$) and two timestamps after (${i+1}, {i+2}$) the chosen pivot timestamp. By doing this, we can ensure that any piecewise B\'ezier curve is at least cubic (more flexible than quadratic~\cite{okumura2012optimization}). The iterative process is stopped when desired number of pivot timestamps are chosen or there remains no possible pivot timestamps, whichever occurs earlier. All the remaining points serve as control points. 	

Now, the least (highest) pivot timestamp may not always correspond to the first (last) frame of the video. Nevertheless for the composite curve to include all the video timestamps, a pivot is necessary at the first and last frame. So, the set of pivot timestamps is modified as $PT_{fin}=PT\cup \{0,F-1\}$. But, the actual 4D coordinates of them cannot be relied upon because their temporal cohesiveness score did not qualify them to be pivots. So, $C_0$ and $C_{F-1}$ is kept same as its nearest pivot. The modified set of points is thus: 
\begin{equation}
	\begin{split}
		C_0 &= \left( x_{min\left( PT\right) },y_{min\left( PT\right) },d_{min\left( PT\right) }\right) \\
		C_1 &= (x_1,y_1,d_1)\\
		\cdots & \;\;\;\;\;\;\;\;\; \cdots\\
		C_{F-2} &= (x_{F-2},y_{F-2},d_{F-2})\\
		C_{F-1} &= \left( x_{max\left( PT\right) },y_{max\left( PT\right) },d_{max\left( PT\right) }\right) 
	\end{split}
\end{equation}

\subsubsection{PolyB\'ezier Curve-Fitting}
\label{polyBezierFit}
Between each pair of consecutive pivot timestamps $i$ and $j$ ($i,j\in PT_{fin}$), a separate ($j-i$)-degree B\'ezier curve is fitted. For such a B\'ezier curve, $C_i$ and $C_j$ serve as end-points and intermediate points $\{C_{i+\delta}\}_{\forall \delta=1,2,\cdots,(j-i-1)}$ serve as control points. Interpolated points will be $(sx_{i+\delta}, sy_{i+\delta}, sd_{i+\delta}, {i+\delta})$ where $(sx_{i+\delta}, sy_{i+\delta}, sd_{i+\delta})=B(\frac{\delta}{j-i})$ for $\delta=0,1,\cdots,(j-i)$. The composite PolyB\'ezier curve thus passes through each of the pivot points, as evident from bottom row of Figure~\ref{fig:bezierInterpolation}. Finally, corrected square-shaped action localization patch at $k^{\text{th}}$ frame ($k=0,1,\cdots ,F-1$) is centered at $(sx_{k}, sy_{k})$, each side being $sd_{k}$. The action-localized video crop is constructed by stitching action localization patches of consecutive frames, in order. 

\section{Experimental Results and Discussion}
\label{exptSec}

\subsection{Datasets and Metric Used}
\label{datasets}
For pre-training the 3D CNNs, we have relied upon some of the recent large-scale video datasets. The datasets used for this purpose are Kinetics-700~\cite{carreira2019short} (700 classes), Moments in Time~\cite{monfort2019moments} (339 classes) and STAIR Action~\cite{yoshikawa2018stair} (100 classes). While the first two datasets mainly contain video collected from YouTube, the latter one contains videos captured and submitted by consumers. We test our models by pre-training on these datasets individually and also on combinations of these, with the methods adopted by Kataoka et al.~\cite{kataoka2020would}. In total, these three datasets combinedly contain about 2 million videos.

For fine-tuning, we have relied upon three commonly-used video action recognition datasets, viz. UCF-101~\cite{soomro2012ucf101}, HMDB-51~\cite{kuehne2011hmdb} and ActivityNet~\cite{caba2015activitynet}. It has been many years now since the first two datasets were introduced and they have become standard benchmarks now. ActivityNet is relatively recent and larger. 

\textbf{UCF-101~\cite{soomro2012ucf101}} consists of 13,320 temporally-trimmed video clips from YouTube, spread over 101 action categories. Videos were of short duration with an average clip-length of 7.21 sec and they comprised realistic human-action scenes.

\textbf{Human Motion Database (HMDB-51)~\cite{kuehne2011hmdb}} comprises of 7,000 temporally-trimmed movie clips. This dataset is challenging mainly because of the diverse human-actions and the poor video quality and stabilization issues.

\textbf{ActivityNet v1.3~\cite{caba2015activitynet}} comprises of about 28,000 untrimmed video clips collected from YouTube, spread over 200 activity categories. The total video duration is almost 850 hours.

For the classification performance, we evaluate accuracy by two metrics: \textit{clip accuracy} i.e. the fraction of correctly classified 16-frame clips in the test-partition and \textit{top-1 accuracy} i.e. the video-level accuracy obtained after combining predictions of all individual constituent 16-frame clips from a test-partition video.

\subsection{Implementation Aspects}
\label{implementation}
Implemented language is Python 3.6.9, its associated computer vision libraries and PyTorch 1.8 for deep learning. They were executed in Ubuntu 16.04 on Intel Core i7-7700K processor. The GPU hardware was Nvidia Titan RTX with 24220MiB total memory. 

For action localization in Section~\ref{axnLoc}, value of $A_{min}$ can be 20-30\% of the frame area. For Section~\ref{pivotSelect}, it was experimentally observed that temporal consistency is maintained best when number of pivots is in [$0.1\times F$, $0.2\times F$], $F$ being the total number of frames in the untrimmed video. To generate training data for the 3D CNN, we uniformly sample a temporal position and select a 16-frame clip around it. Thereafter for spatial sampling, we randomly select one of the four corners of the video frame or center and perform multi-scale cropping from one of $\left\lbrace 1, \frac{1}{2^{\frac{1}{4}}}, \frac{1}{2^{\frac{1}{2}}}, \frac{1}{2^{\frac{3}{4}}}, \frac{1}{2} \right\rbrace $, similar in lines of Hara et al.~\cite{hara2018can}. We experiment with spatial cropping of $112\times 112$ (whereby, each sample spans $3\times16\times112\times112$) and $56\times 56$ (whereby, each sample spans $3\times16\times56\times56$). 

\subsection{Comparison with State-of-the-Arts}
\label{comparison}
In Table~\ref{tab_sota}, we tabulate the top-1 accuracy obtained by recent state-of-the-arts that employ 3D CNN or 2D CNN with LSTM~\cite{jana2019key}, along with the input clip dimension (represented as \#Channels $\times$ \#Frames $\times$ FrameHeight $\times$ FrameWidth). One one hand, it can be concluded that pre-trained 3D CNN architectures outperforms their relatively complex 2D counterparts combined with LSTM, signifying the potential of 3D CNNs in video classification. Alongside it is interesting to note that for the same 3D CNN architecture, larger input dimension is seen to proffer higher accuracy (e.g. I3D in Table~\ref{tab_sota}). This can be attributed to the fact that in all the three datasets the original video size is larger on an average ($240\times320$ in UCF-101, $240\times\_$ in HMDB-51 and variable in ActivityNet v1.3) -- thus, downsizing and random cropping leads to poor results. 

\begin{table*}[!h]
	\centering
	\caption{Performance of recent state-of-the-art methods using 3D CNN or 2D CNN + LSTM architectures on the three datasets, in terms of top-1 accuracy (\%).}
	\resizebox{1\textwidth}{!}{%
		\begin{tabular}{c|c||c|c|c}
			\hline
			\textbf{Method} & \textbf{Input Dimension} &
			\textbf{UCF-101~\cite{soomro2012ucf101}} & 
			\textbf{HMDB-51~\cite{kuehne2011hmdb}} &
			\textbf{ActivityNet v1.3~\cite{caba2015activitynet}}%
			\TstrutNorm\BstrutNorm\\ \hline
			3D-ResNet34~\cite{kataoka2020would} & $3\times16\times112\times112$ & 88.8  & 59.5 & 71.4\TstrutNorm\\
			3D-ResNet200~\cite{kataoka2020would} & $3\times16\times112\times112$ & 92.0  & 68.1 & 77.8\\
			(2+1)D-ResNet50~\cite{tran2018closer} & $3\times16\times112\times112$ & 91.2  & 66.4 & 74.0\\
			(2+1)D-ResNet200~\cite{tran2018closer} & $3\times16\times112\times112$ & 79.5  & 52.9 & 58.9\\
			I3D
			(RGB+Flow)~\cite{carreira2017quo} & $3\times64\times224\times224$ & 98.0 & 80.7 & -\\
			I3D (RGB+Flow)~\cite{carreira2017quo} & $3\times64\times112\times112$ & 93.2  & 70.5 & -\\
			3D-ResNeXt101~\cite{xie2017aggregated} & $3\times64\times112\times112$ & 94.5  & 70.2 & -\\
			3D-ResNeXt101~\cite{xie2017aggregated} & $3\times16\times112\times112$ & 90.7  & 63.8 & -\\
			FAST3D-DenseNet121~\cite{stergiou2019spatio} & $3\times24\times224\times224$ & 89.5  & 55.4 & -\\			
			2D-ResNet50 + LSTM~\cite{jana2019multi} & $3\times10\text{(Key-Frm)}\times224\times224$ & 89.0  & 61.9 & -\\
			2D-Inception + LSTM (YT-8M)~\cite{abu2016youtube} & $3\times20\text{(Random)}\times224\times224$ & -  & - & 75.6\\
			LiteEval (Course2Fine CNN+LSTM)~\cite{liteeval2019wu} & $3\times25\text{(Random)}\times224\,\&112\,\times224\,\&\,112$ & -  & - & 72.7\BstrutNorm\\
			\hline
	\end{tabular}}
	
	\label{tab_sota}
\end{table*}

\begin{table*}[!h]
	\centering
	\caption{Accuracy obtained on the three datasets used for fine-tuning, after different combinations of pre-training on the 3D-ResNet50 model. Here, RC represents random cropping performed on input videos and AL denotes application of the proposed action localization on input videos. $112\times 112$ or $56\times 56$ denote the spatial dimensions. Values in each cell is of the form [Clip Accuracy (\%) / Top-1 Accuracy (\%)].}
	\resizebox{\textwidth}{!}{%
		\begin{tabular}{cc||c|c|c||c|c|c||c|c|c}
			\hline
			&
			\multirow{11}{*}{\backslashbox{\textbf{Pre-Training}}{\textbf{Fine-Tuning}}} &
			\multicolumn{3}{c||}{\textbf{UCF-101~\cite{soomro2012ucf101}}} &
			
			\multicolumn{3}{c||}{\textbf{HMDB-51~\cite{kuehne2011hmdb}}} & 
			
			\multicolumn{3}{c}{\textbf{ActivityNet v1.3~\cite{caba2015activitynet}}}\TstrutNorm\\
			& & \multicolumn{3}{c||}{Clip Accuracy (\%) / Top-1 Accuracy (\%)} & \multicolumn{3}{c||}{Clip Accuracy (\%) / Top-1 Accuracy (\%)} & \multicolumn{3}{c}{Clip Accuracy (\%) / Top-1 Accuracy (\%)} \TstrutNorm\BstrutNorm\\\cline{3-11}
			
			& & 
			\begin{tabular}{@{}c@{}}RC\\ 
				\multirow{1}{*}{\rotatebox[origin=c]{30}{$112\times 112$}}\end{tabular} & 
			\begin{tabular}{@{}c@{}}RC \\ \multirow{1}{*}{\rotatebox[origin=c]{30}{$56\times 56$}}\end{tabular} & 
			\begin{tabular}{@{}c@{}}AL \\
				
				\multirow{1}{*}{\rotatebox[origin=c]{30}{$56\times 56$}}\end{tabular} & 
			\begin{tabular}{@{}c@{}}RC \\ \multirow{1}{*}{\rotatebox[origin=c]{30}{$112\times 112$}}\end{tabular} & 
			\begin{tabular}{@{}c@{}}RC \\ \multirow{1}{*}{\rotatebox[origin=c]{30}{$56\times 56$}}\end{tabular} & 
			\begin{tabular}{@{}c@{}}AL \\
				
				\multirow{1}{*}{\rotatebox[origin=c]{30}{$56\times 56$}}\end{tabular} & 
			\begin{tabular}{@{}c@{}}RC \\ \multirow{1}{*}{\rotatebox[origin=c]{30}{$112\times 112$}}\end{tabular} & 
			\begin{tabular}{@{}c@{}}RC \\ \multirow{1}{*}{\rotatebox[origin=c]{30}{$56\times 56$}}\end{tabular} & 
			\begin{tabular}{@{}c@{}}AL \\ \multirow{1}{*}{\rotatebox[origin=c]{30}{$56\times 56$}}\end{tabular}\Tstrut\Bstrut\\\hline\hline

			& \textbf{Kinetics-700~\cite{carreira2019short} (K)} & 87.9 / 92.0 & 79.9 / 83.7 & 87.0 / 91.5 & 57.1 / 66.0 & 44.7 / 52.8 & 59.0 / 65.5 & 56.1 / 75.9 & 49.1 / 67.2 & 57.5 / 74.3 \TstrutNorm\\
			& \textbf{Moments in Time~\cite{monfort2019moments} (M)} & 80.5 / 85.5 & 71.1 / 76.5 & 81.0 / 86.3 & 52.7 / 62.6 & 48.9 / 53.1 & 54.3 / 63.1 &  46.4 / 65.9 & 35.7 / 53.4 & 46.5 / 66.2 \\
			& \textbf{STAIR Action~\cite{yoshikawa2018stair} (S)} & 50.7 / 55.6  & 40.3 / 43.1 & 50.1 / 55.2 &  26.8 / 31.2 & 17.3 / 20.8 & 26.7 / 32.0 & 24.2 / 36.5 & 16.9 / 24.3 & 24.9 / 34.6 \BstrutNorm\\ \hline

			& \textbf{K + M} & 89.1 / 92.9 & 82.1 / 84.3 & 88.2 / 90.3 & 60.4 / 69.4 & 48.2 / 58.9 & 61.3 / 69.1 & 57.4 / 77.0 & 50.4 / 68.3 & 57.1 / 76.2 \TstrutNorm\\
			& \textbf{M + S} & 76.4 / 81.3 & 66.2 / 70.7 & 76.9 / 82.6 & 48.9 / 56.4 & 43.4 / 50.4 & 48.3 / 55.9 & 43.5 / 62.8 & 36.3 / 52.3 & 45.7 / 63.1 \\ 
			& \textbf{K + S} & 87.1 / 91.0 & 77.4 / 80.8 & 86.4 / 89.9 & 57.0 / 64.9 & 49.1 / 55.6 & 56.4 / 62.1 & 56.0 / 74.9 & 49.0 / 65.2 & 55.4 / 74.2 \\
			& \textbf{K + M + S} & 88.3 / 92.3 & 79.0 / 84.1 & 87.8 / 91.3 & 58.8 / 67.8 & 52.3 / 60.0 & 60.1 / 67.2 & 56.6 / 75.8 & 50.3 / 65.3 & 53.9 / 73.7 \BstrutNorm\\ 
			\hline
	\end{tabular}}
	
	\label{tab_preFineTune}
\end{table*}

To evaluate the proposed action localization method, we study the clip and top-1 accuracy on 3D-ResNet50 architecture for three cases in Table~\ref{tab_preFineTune}. The \textit{first case} (denoted as RC $112\times 112$ in Table~\ref{tab_preFineTune}) is the baseline on the raw video as in Kataoka et al.~\cite{kataoka2020would}. That is, training is done on a random temporally-cropped 16-frame video clip, involving random multi-scale sized $\times10$ augmented frames by four-corner or center cropping, resized to $112\times112$. The \textit{second case} (denoted as RC $56\times 56$ in Table~\ref{tab_preFineTune}) is similar to first, only exception being that the final spatial crop-size is $56\times56$. For the \textit{third case} (denoted as AL $56\times 56$ in Table~\ref{tab_preFineTune}), instead of inputting the raw video we take the modified video obtained by stitching the temporally consistent action localized patches obtained by the proposed method. The final spatial crop-size is $56\times56$, keeping in mind that subject of a video typically covers a minor portion of the whole frame. In terms of outcomes, firstly there is a huge gain in GPU memory usage when using smaller spatial dimensions. As for example, training 3D-ResNet50 with batch-size $128$ and input-size $3\times16\times112\times112$ consumes $\sim$21388 MB, while it is only $\sim$6438 MB for $3\times16\times56\times56$. Secondly, it can be observed that results in the \textit{third case} (AL $56\times56$) of applying proposed action localization cropping is always better than random-crop of same dimension in the \textit{second case} (RC $56\times56$). Furthermore, sometimes it even outperforms instances of higher spatial dimension with random-crop in the \textit{first case} (RC $112\times112$). 

\section{Conclusion and Future Scope}
\label{conclusion}
In this paper, we study the impact of deploying a video-specific cropping technique based on localizing the actions, which in turn, makes it much more focussed and targeted towards the video subject. The proposed method plays a key role in highlighting the video subject, while eradicating insignificant background information. To achieve the dynamic cropping, we start with identifying a single motion-heavy patch from each video frame. Temporal consistency is maintained amongst these square-shaped patches by choosing few pivot points and fitting a 3D polyB\'ezier curve. We assess the proposed method in the task of event classification from videos, possessing large-scale content diversity, by feeding dynamically cropped videos to various 3D CNN frameworks. In terms of the classification accuracy achieved on UCF-101, HMDB-51 and ActivityNet v1.3 datasets, our approach is seen to outperform the results on state-of-the-art 3D CNNs that employ random- or center cropping techniques. The results reinforce our idea of a more semantic and dynamic cropping approach than an intuitively random one. Efficient cropping to a low input size leaves enough free GPU memory that permits easy actuation of other add-on tasks that use 3D CNNs as a backbone e.g., temporal action localization. Also, the proposed solution framework or dynamic cropping techniques in general, hints at having telling effects on other computer vision tasks, such as the likes of video segmentation, 3D object detection etc. which we hope to pursue in our future scope of work.

	%
%
%
%
\bibliographystyle{IEEEtran}
\bibliography{IEEEabrv,bibFile}

\end{document}